\documentclass{article}

\usepackage[final]{corl_2018} 

\usepackage{xspace}
\usepackage{amsfonts}       
\usepackage{nicefrac}       
\usepackage{microtype}      
\usepackage{graphicx}
\usepackage{color}
\usepackage{amssymb,amsmath,bm,amsthm}
\usepackage{subcaption}
\usepackage[usenames,dvipsnames]{xcolor}  
\usepackage{multirow}

\usepackage{algorithm}
\usepackage{algpseudocode}  
\usepackage[]{todonotes}  
\usepackage{tabulary}
\usepackage{arydshln}

\usepackage{sidecap}
\usepackage{subcaption}
\usepackage{wrapfig}

\usepackage{enumitem}
\setlist[itemize]{nosep}  

\usepackage{titlesec}
\titlespacing*{\section}
{0pt}{.4\baselineskip}{0.5\baselineskip}
\titlespacing*{\subsection}
  {0pt}{0.25\baselineskip}{0.2\baselineskip}

\newcommand{\seclabel}[1]{\label{sec:#1}}

\newcommand{\tabref}[1]{Table~\ref{tab:#1}}
\newcommand{\Sys}{\textsc{RoboTurk}\xspace}
\newcommand{\sysAbbr}{\textsc{RoboTurk}\xspace}

\makeindex             

\title{\sysAbbr: A Crowdsourcing Platform for \\ Robotic Skill Learning through Imitation}

\author{
Ajay Mandlekar$^\ddagger$,
Yuke Zhu,
Animesh Garg,
Jonathan Booher,
Max Spero,
Albert Tung, \\ \\
\textbf{Julian Gao,
John Emmons,
Anchit Gupta,
Emre Orbay,
Silvio Savarese,
Li Fei-Fei}
\\ \\
Department of Computer Science, Stanford University\\
$^\ddagger$\texttt{amandlek@stanford.edu}
}

\begin{document}
\maketitle

\begin{abstract}
Imitation Learning has empowered recent advances in learning robotic manipulation tasks by addressing shortcomings of Reinforcement Learning such as exploration and reward specification. 
However, research in this area has been limited to modest-sized datasets due to the difficulty of collecting large quantities of task demonstrations through existing mechanisms.
This work introduces \sysAbbr to address this challenge. 
\sysAbbr is a crowdsourcing platform for high quality 6-DoF trajectory based teleoperation through the use of widely available mobile devices (e.g. iPhone).
We evaluate \sysAbbr on three manipulation tasks of varying timescales (15-120s) and observe that our user interface is statistically similar to special purpose hardware such as virtual reality controllers in terms of task completion times. 
Furthermore, we observe that poor network conditions, such as low bandwidth and high delay links, do not substantially affect the remote users' ability to perform task demonstrations successfully on \Sys. Lastly, we demonstrate the efficacy of \sysAbbr through the collection of a pilot dataset; using \sysAbbr, we collected 137.5 hours of manipulation data from remote workers, amounting to over 2200 successful task demonstrations in 22 hours of total system usage. We show that the data obtained through \sysAbbr enables policy learning on multi-step manipulation tasks with sparse rewards and that using larger quantities of demonstrations during policy learning provides benefits in terms of both learning consistency and final performance. For additional results, videos, and to download our pilot dataset, visit \href{http://roboturk.stanford.edu/}{\texttt{roboturk.stanford.edu}}
\end{abstract}


\renewcommand{\textfloatsep}{3mm}
\renewcommand{\dbltextfloatsep}{3mm}

\section{Introduction}


Large-scale datasets, consisting of millions of data points, have accelerated performance in computer vision and natural language tasks and enabled many practical applications. Richly annotated data has provided a fertile ground for developing and evaluating a broad set of learning algorithms for problems such as image classification and machine translation~\cite{deng2009imagenet,lin2014microsoft,rajpurkar2018squad2}. 
Supervised data is similarly useful for sequential decision making problems in robotics.


Recent research has successfully employed Reinforcement Learning (RL), Self-Supervised Learning (SSL), and Imitation Learning (IL) for short-horizon skill learning, such as pushing and grasping objects~\cite{yu2016more,levine2016learning,fang2018learning}. However, RL methods require reward specification, face the burden of efficient exploration in large state spaces, and need large amounts of interaction with the environment to collect sufficient data for policy learning.
SSL has been used successfully to collect and learn from large quantities of data in both simulated~\cite{mahler2017dex,kasper2012kit,goldfeder2009columbia} and physical settings~~\cite{levine2016learning,pinto2016supersizing,kalashnikov2018qt} for tasks such as grasping. However, the data has very low signal-to-noise ratio due to random exploration.
In contrast, IL uses expert demonstrations to reduce environment interaction during skill learning. Imitation Learning is commonly posed either as inverse reinforcement learning~\cite{abbeel2011inverse,ng2000algorithms} or behavioral cloning~\cite{argall2009survey,pomerleau1989alvinn}, both of which require demonstrations.
Prior works show that data-driven IL methods improve both the sample efficiency of policy learning~\cite{abbeel2010autonomous,vevcerik2017leveraging,krishnan2016swirl} and the performance of the trained policy~\cite{gupta2016learning,boularias2011relative,ho2016generative}.

There are many conventional mechanisms for collecting task demonstrations for robot learning. One popular option is to kinesthetically guide a robot through a desired trajectory~\cite{argall2009survey}. Although intuitive, this mechanism for supervision can be tedious, and is often limited to collecting tens of demonstrations, while policy learning requires hundreds or even thousands of demonstrations~\cite{osentoski2010crowdsourcing}. Alternatively, teleoperation-based techniques for collecting task demonstrations have been employed for over two decades~\cite{Goldberg94beyondthe,sung2018robobarista,hokayem2006bilateral}. Teleoperated supervision can be provided through contemporary game interfaces such as a keyboard and mouse~\cite{kent2017comparison, leeper2012strategies}, a video game controller~\cite{laskey2017comparing}, a 3D-mouse ~\cite{Dragan-2012-7572,zhu2018reinforcement}, special purpose master-slave interfaces\cite{akgun2012novel,liang2017using}, or through free-space positioning interfaces such as virtual reality (VR)~\cite{whitney2017comparing,zhang2017deep,lipton2018baxter}. 

\begin{figure}[t!]
\centering
  \includegraphics[width=1.\linewidth]{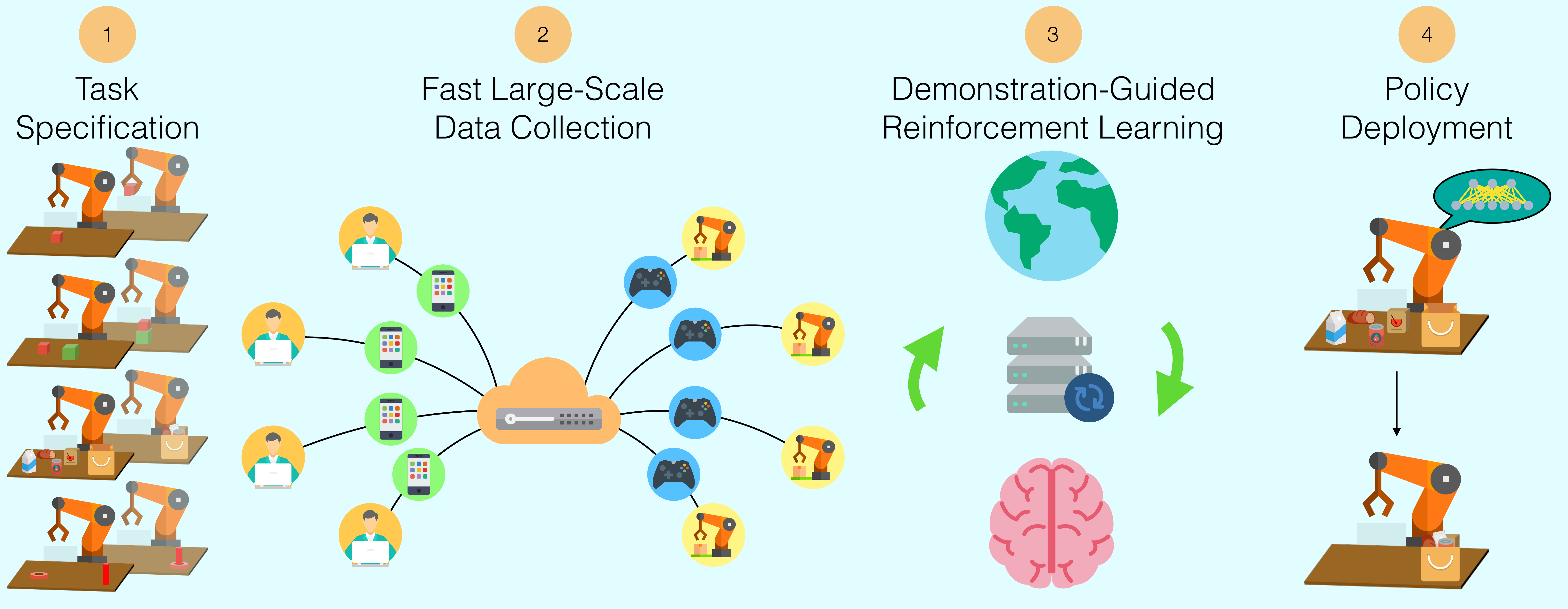}
  \caption{\textbf{System overview of \sysAbbr{}.} \sysAbbr{} enables quick imitation guided skill learning. Our system consists of the following major steps: 1) specifying a task, 2) collecting a large set of task demonstrations using \sysAbbr{}, 3) using demonstration-augmented reinforcement learning to learn a policy, and 4) deploying the learned skill in the domain of interest.}
  \label{fig:sys_overview}
\end{figure}

Game interfaces frequently introduce artifacts in trajectories that reduce the utility of this data for imitation~\cite{akgun2012keyframe}. For instance, due to concurrent control of only a subspace of actions, the trajectories can be longer, exhibit axis-aligned movement, and lack natural variations in motion. By contrast, free-space positioning interfaces such as VR enable the user to directly move the robot's end effector by moving a hand controller in 3D space, enabling fine-grained dexterous control. 
Interestingly, Zhang et al.~\cite{zhang2017deep} showed that simpler algorithms such as variants of behavior cloning can be made to learn short-horizon manipulation tasks with data on the order of a few hundred VR demonstrations. 
However, the requirement of special purpose VR hardware and client-side compute resources has limited the deployment of these interfaces on crowdsourcing platforms such as Amazon Mechanical Turk where a typical worker is more likely to have a smartphone than VR hardware~\cite{superdata2017vr,apple2017report}. 

In other domains, large-scale supervision for datasets is often collected with the assistance of crowdsourcing~\cite{deng2009imagenet,rajpurkar2018squad2}. This enables a scalable mechanism for diverse human supervision on an extensive set of problem instances. However, collecting large amounts of data has been a challenge for continuous control tasks, as they demand real-time interaction and feedback from annotators, placing difficult constraints on remote teleoperation platforms. Data collection mechanisms for skill learning in robotics need demonstrations from remote users that are both \textit{natural} (which game interfaces tend to lack) and \textit{plentiful} (which free-space positioning interfaces tend to lack).

This paper proposes \Sys, a platform to address the challenge of collecting large, crowdsourced sets of task demonstrations for robot manipulation tasks. 
\sysAbbr enables: (a) real-time control of simulated systems, (b) accessibility to a large population of users, and (c) simultaneous platform use by many concurrent users. \sysAbbr is one of the first platforms to facilitate crowdsourcing of trajectory-level supervision.

\noindent \textbf{Contributions.} The main contributions of this paper are:
\begin{enumerate}[
    topsep=0pt,
    leftmargin=*,
    ]
    
\item We introduce \sysAbbr{}, a data collection platform that allows remote workers to log on to a website and collect task demonstrations using a smartphone as a motion controller. \sysAbbr is supported by a cloud-based simulation backend that streams video to a client's web browser using low-latency communication protocols. This ensures homogeneous quality of service regardless of a client's compute resources, resulting in a platform that is intuitive to use and has a low barrier to entry -- the core requirements of a crowdsourced task. \sysAbbr{} supports multiple robots, tasks, and simulators, and can easily be extended to support others. 
\item We conduct a user study comparing the performance of three other interfaces, including a VR interface, on simulated lifting and bin picking tasks. Our platform is similar to VR in performance and better than 3D-mouse and keyboard interfaces. We also evaluate the performance of \sysAbbr under poor network conditions using a network simulator.
\item We present an initial dataset consisting of over 2200 task demonstrations, amounting to 137 hours of data collected in 20 hours of system usage with contracted workers. 
\item We show that the data collected through \sysAbbr{} enables policy learning on challenging manipulation tasks with sparse rewards and that using larger quantities of demonstrations during policy learning provides benefits in terms of both learning consistency and final performance.
\end{enumerate}

\begin{figure}[t]
\centering
  \includegraphics[width=1.\linewidth]{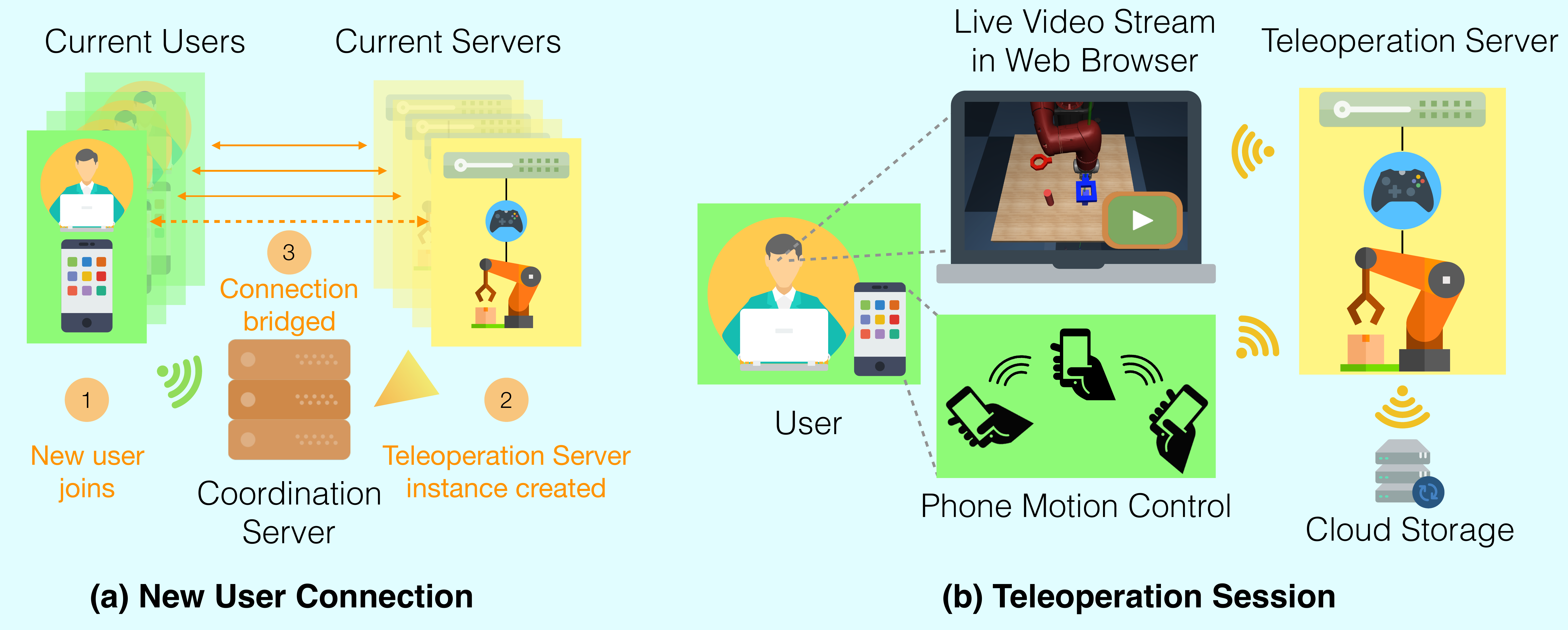}
  \caption{\textbf{System diagram of \sysAbbr{}.} A new user connects to a website to join the system, and a coordination server launches a dedicated teleoperation server for the user, as shown in \textbf{(a)}. The coordination server then establishes direct communication channels between the user's web browser and iPhone and the teleoperation server to start the teleoperation session. The user controls the simulated robot by moving their phone, and receives a video stream as feedback in their web browser, as shown in \textbf{(b)}. After every successful demonstration, the teleoperation server pushes the collected data to a cloud storage system.}
  \label{fig:sys_diagram}
\end{figure}

\section{\sysAbbr: Design and Development of the Proposed Platform}

We present \sysAbbr, a cloud-based large-scale data collection platform for robot learning tasks that enables the collection of thousands of task demonstrations within days. 


\subsection{Platform Features} \label{sec:design_reqs}


\begin{enumerate}[
    topsep=0pt,
    leftmargin=*,
    itemindent=3ex
]
\item \textbf{Real-time control of simulated robots.} \sysAbbr enables users to control robots in simulated domains and provides users with real-time video and haptic feedback.
\item \textbf{Accessible to a large population of users.} In order to enable large-scale data collection, \sysAbbr is easily accessible to the typical workers on crowdsourcing platforms such as Amazon MTurk. \sysAbbr limits the hardware requirements necessary to use it, minimizes the complexity of any client-side application users interact with, and makes its internal communication protocols agnostic to the geographic proximity of users. 
\item \textbf{Capable of providing real-time feedback and low-latency robot control for many simultaneous users.} In order to collect thousands of robot demonstrations quickly, \sysAbbr is capable of interacting with many users at the same time. This requires careful system design with respect to user session management, communication protocols used to connect users to simulator instances, and the quality of the cloud machines used to host the platform.
\item \textbf{Modular design to enable platform extensibility.} \sysAbbr adopts a modular design paradigm to decouple task and controller design from the rest of the infrastructure --- this makes it easy to extend \sysAbbr to new tasks, physical simulators, and robots.
\end{enumerate}

\subsection{Platform Design and Implementation}
\sysAbbr{} implements the features listed in Sec.~\ref{sec:design_reqs} by deploying its core components on cloud infrastructure. \sysAbbr{} offloads robot simulation to powerful cloud machines instead of local execution. The robot simulation machines communicate over low-latency channels with users to receive commands controlling the position of the simulated robot arms; simulator camera views of the robots are rendered in real-time and transmitted to the users over separate low-latency links. \sysAbbr{} leverages Web Real-Time Communication (WebRTC), an open-source framework for low-latency transmission of both robot control commands and the rendered camera views from the robot simulator. Fig.~\ref{fig:sys_diagram} outlines how system components are involved in (1) letting a new user seamlessly connect to the platform and (2) allowing a user to collect demonstrations inside a dedicated teleoperation session.

By taking this design approach, \sysAbbr{} eliminates the need for users to have a powerful local workstation and to install or configure any software. Instead, users only need to have (1) an iPhone that is compatible with Apple's ARKit framework (iPhone 6S and above) and (2) access to a separate device with a web browser, both of which are ubiquitous \cite{apple2017report}. 


\noindent \textbf{User Endpoint.}
The user receives a real-time video stream of the robot arm in their browser window and moves their iPhone to control the robot arm. The phone's pose is tracked via Apple's ARKit platform, which uses a combination of camera frames and motion sensor data. The phone's pose is packaged along with some status information and transmitted through our platform to a teleoperation server instance, which is a dedicated process responsible for controlling the robot. The teleoperation server sends back task-specific information, allowing the user to receive haptic feedback through phone vibrations when the robot arm makes contact with objects in the simulation environment.

\noindent \textbf{Coordination Server.}
The coordination server is responsible for creating and maintaining user teleoperation sessions. As Fig.~\ref{fig:sys_diagram}a shows, when a new user enters the system, the coordination server spawns a dedicated teleoperation server instance, so that every user can control their own simulated robot arm. The coordination server then establishes two low latency WebRTC communication channels --- one between the user's browser and the teleoperation server and another between the user's phone and the teleoperation server. 

\noindent \textbf{Teleoperation Server.}
Each teleoperation server receives phone commands from exactly one user and processes them to control the robot arm in the simulator. Phone poses are mapped to a set of joint velocities to control the simulated robot at a regulated rate. Additional details in Appendix~\ref{sec:sysDetails}.

\begin{figure}[t]
\centering
\includegraphics[width=1.0\linewidth]{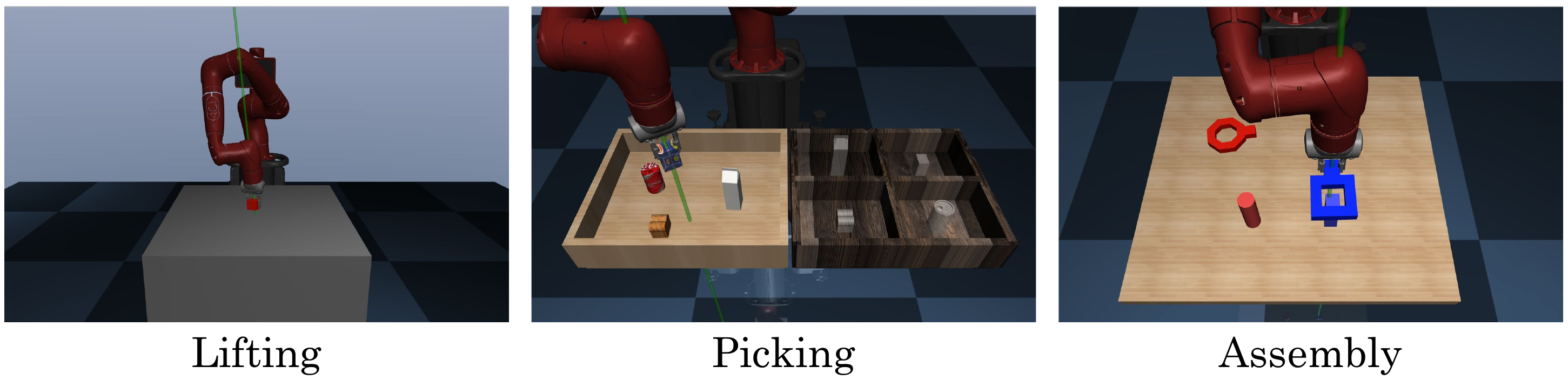}
\caption{\textbf{Tasks}: We evaluated our platform on three simulated manipulation tasks which contain a 7-DoF Sawyer robot arm and various objects. In the \textit{lifting} task (left) the objective is to control the robot arm to grab the cube and lift it. In the \textit{picking} task (middle) the objective is to place each object into its corresponding bin. In the \textit{assembly} task (right), the objective is to place a round nut and a square nut onto their corresponding pegs.}
\label{fig:task-diagram}
\end{figure}

\section{System Analysis}
\label{sec:analysis}

\begin{figure}[!t]
    \centering
    \begin{minipage}[c]{.28\linewidth}
        \centering
            \includegraphics[width=\linewidth]{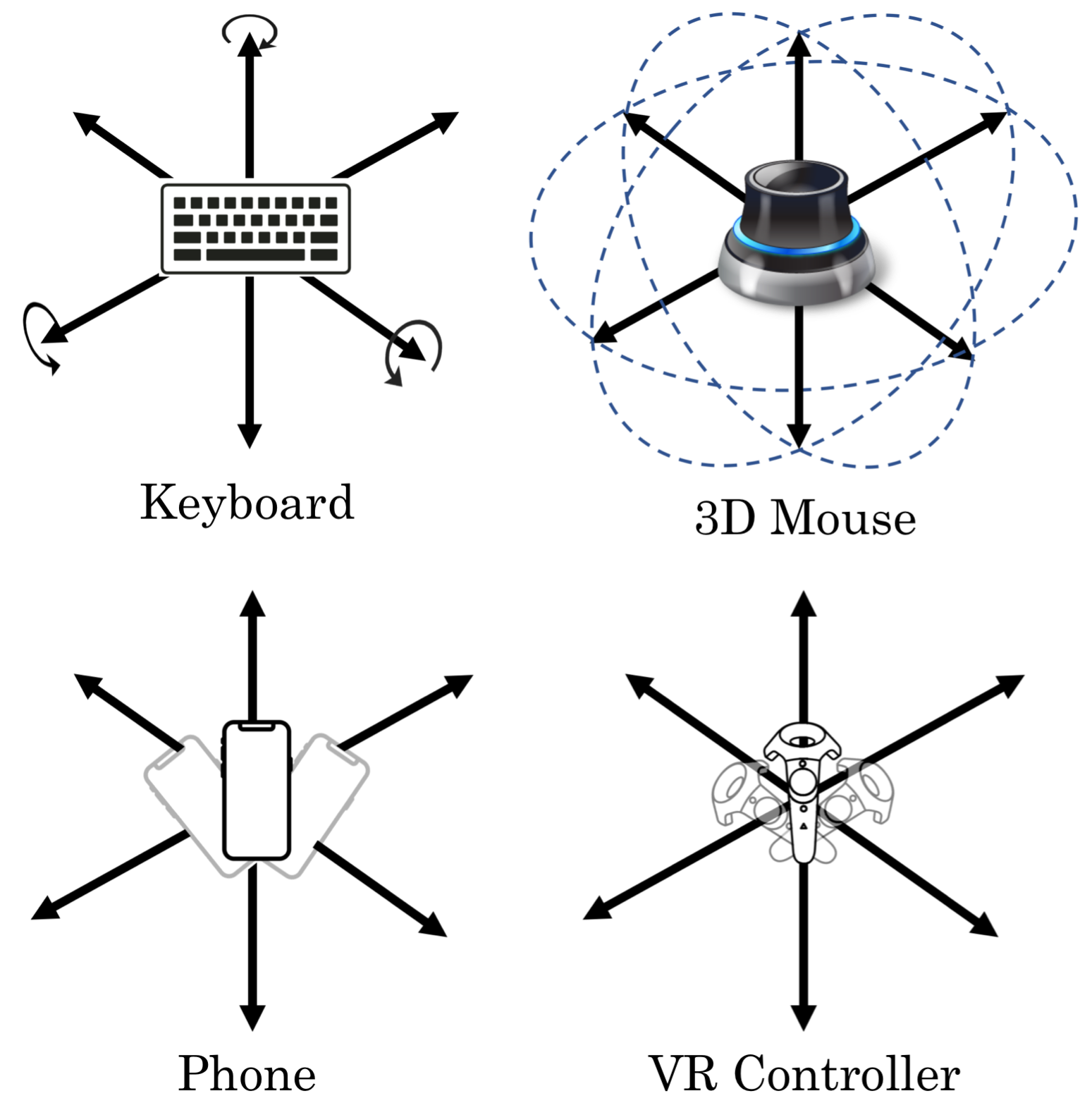}
    \end{minipage}
    \begin{minipage}[c]{.68\linewidth}
    \resizebox{0.98\linewidth}{!}{
        \begin{tabular}{l|c|c|c|c|}
            \cline{2-5}
                & Keyboard & 3D Mouse & VR Controller & Phone \\ \hline
            \multicolumn{1}{|l|}{\multirow{2}{*}{Keyboard}} & \multirow{2}{*}{-} & 0.575 & 0.900 & 0.725 \\ 
            \multicolumn{1}{|l|}{} & & (0.000) & (0.000)  & (0.000) \\ \hline
            \multicolumn{1}{|l|}{\multirow{2}{*}{3D Mouse}} & 0.575 & \multirow{2}{*}{-} & 0.375 & 0.325 \\
            \multicolumn{1}{|l|}{} & (0.000) & & (0.005)  & (0.022) \\ \hline
            \multicolumn{1}{|l|}{\multirow{2}{*}{VR Controller}} & 0.900 & 0.375 & \multirow{2}{*}{-} & \textbf{0.225} \\ 
            \multicolumn{1}{|l|}{} & (0.000) & (0.005) & & \textbf{(0.231)} \\ \hline
            \multicolumn{1}{|l|}{\multirow{2}{*}{Phone (ours)}} & 0.725 & 0.325 & \textbf{0.225} & \multirow{2}{*}{-} \\ 
            \multicolumn{1}{|l|}{} & (0.000) & (0.022) & \textbf{(0.231)} & \\ \hline
        \end{tabular}
    }
    \end{minipage}
    \caption{\textbf{UI Comparison:} \textbf{(a)} Illustration of Interfaces and the movement they allow --- axis-aligned (Keyboard), 6-DoF (3D Mouse), and Free-Space (VR and Phone). \textbf{(b)} Table: Kolmogorov-Smirnov statistics between distributions for completion times in the \textit{picking} task across different interfaces, followed by associated $p$-values. Pairwise K-S statistic is used as a measure of difference between underlying distributions.
    Based on statistical significance level of 5\%, we observe that the completion times follow the order: Phone $\approx$ VR Controller $\succ$ 3D Mouse $\succ$ Keyboard}
\label{fig:ui_diags}
\end{figure}

In this section, we benchmark \sysAbbr and evaluate some of our design choices. In particular, we evaluate our iPhone interface against other common interfaces motivated by prior work~\cite{leeper2012strategies, zhu2018reinforcement, zhang2017deep} and the performance of our platform under low bandwidth and high delay network conditions. 

\subsection{Tasks}

Our experiments use three simulated tasks: \textit{Block Lifting} (\textit{lifting}), \textit{Bin Picking} (\textit{picking}), and \textit{Nut-and-peg Assembly} (\textit{assembly}), as illustrated in Fig. \ref{fig:task-diagram}. \textit{Lifting}, a simple task where a Sawyer robot arm must lift a cube, serves as a diagnostic example, while the \textit{picking} and \textit{assembly} tasks, which consist of sorting objects into bins and fitting nuts onto pegs respectively, are more challenging settings. These are a part of the \textsc{Surreal} Robotics Suite \cite{fan2018surreal}, a set of manipulation tasks developed using the MuJoCo physics engine \cite{todorov2012mujoco} inspired by the tasks in World Robot Summit~\cite{WorldRobotSummit}. Please refer to Appendix~\ref{sec:taskDetails} for task details.

To evaluate our platform, we conducted a user study with 8 university students aged 18-30. Each user provided 20 demonstrations on the \textit{lifting} task and 5 on the \textit{picking} task for each of 8 different test conditions. In the first 4 test conditions, we varied the control interfaces. In the latter 4 test conditions, we varied the network conditions that users experienced while controlling the robot arm. 

\subsection{User Interface Evaluation}

As Fig.~\ref{fig:ui_diags}a shows, we compare four user interfaces for robot control: Keyboard, 3D Mouse, Virtual Reality (VR) Controller, and Phone (ours). The HTC Vive is special-purpose VR hardware that uses external-tracked pose for hand-held controllers, which we map one-to-one to the robot end effector pose. The tracking is high fidelity, but every remote teleoperator would require access to specialized VR hardware. The 3D mouse interface is a joystick that can translate simultaneously along x, y, and z and rotate about these axes as well. The local movements of the 3D mouse are mapped to the robot end effector. While it may not offer the same accuracy as the HTC Vive, it is only a fraction of the cost, and can be seen as a compromise between cost and performance. The keyboard is the most ubiquitous of the four interfaces considered, but it lacks natural 3D control and could easily lead to degenerate, axis-aligned demonstrations that do not resemble free-form, 6-DoF movement. In our implementation, we used different keys to translate or rotate by a fixed amount in each direction, and a separate key to open and close the gripper. An ideal control interface would be as commonplace as a keyboard while as intuitive as a VR controller. 

We designed the user interface on the phone to try and combine the ease of control and accuracy of the VR controller with the ubiquity of the keyboard.
Pose-tracking on phones has been massively improved in the last few years, especially with the advent of ARKit and ARCore platforms. We utilize ARKit to track the phone's pose using the camera and internal gyroscope. While the phone's tracking mechanism is often noisy and does not match the fidelity of the VR controllers, we observed that our subjects often adapted to account for tracking noise after collecting a few demonstrations and getting familiar with the interface.

\begin{figure}[t]
\centering
\includegraphics[width=.475\linewidth]{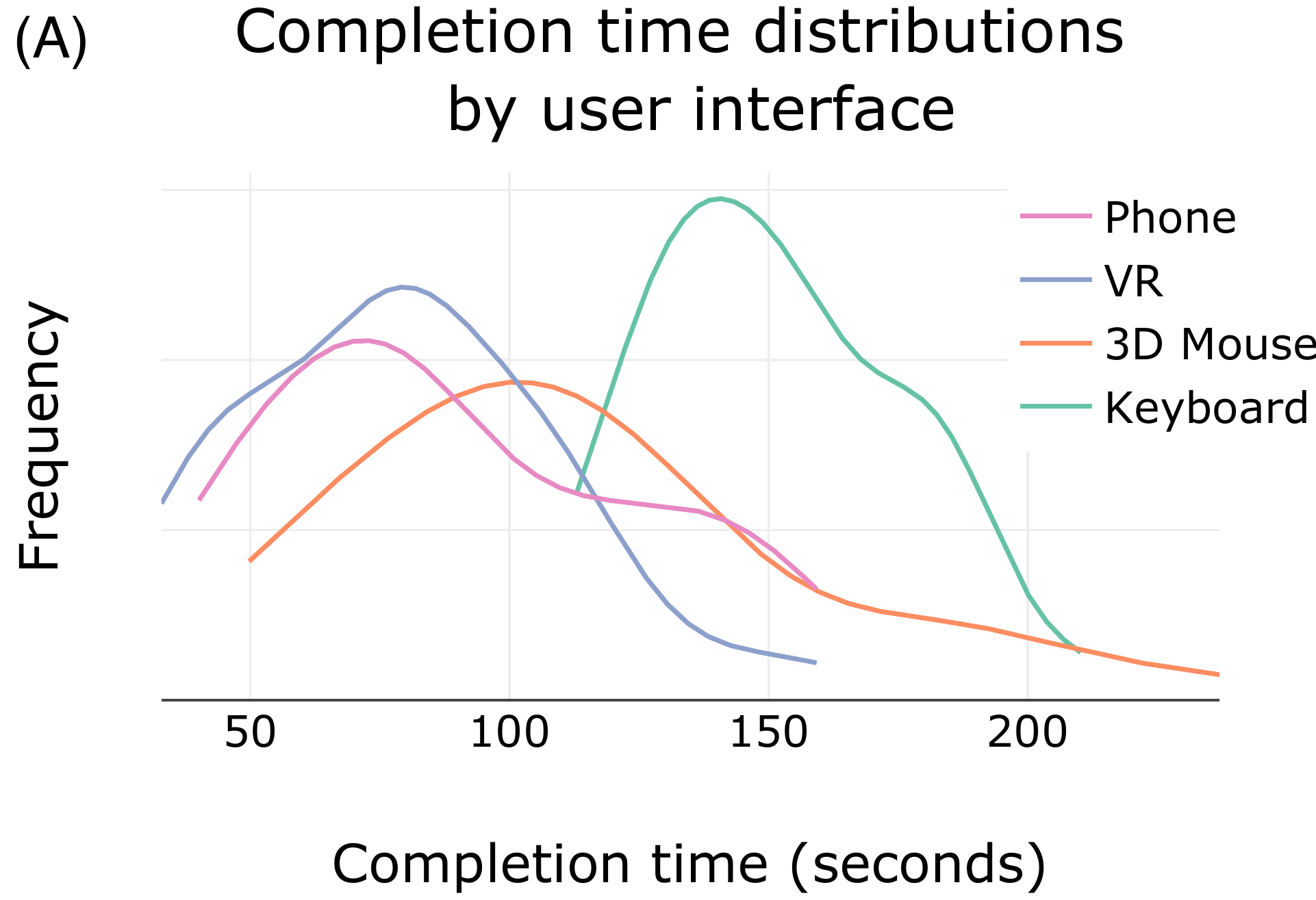}~
\includegraphics[width=.495\linewidth]{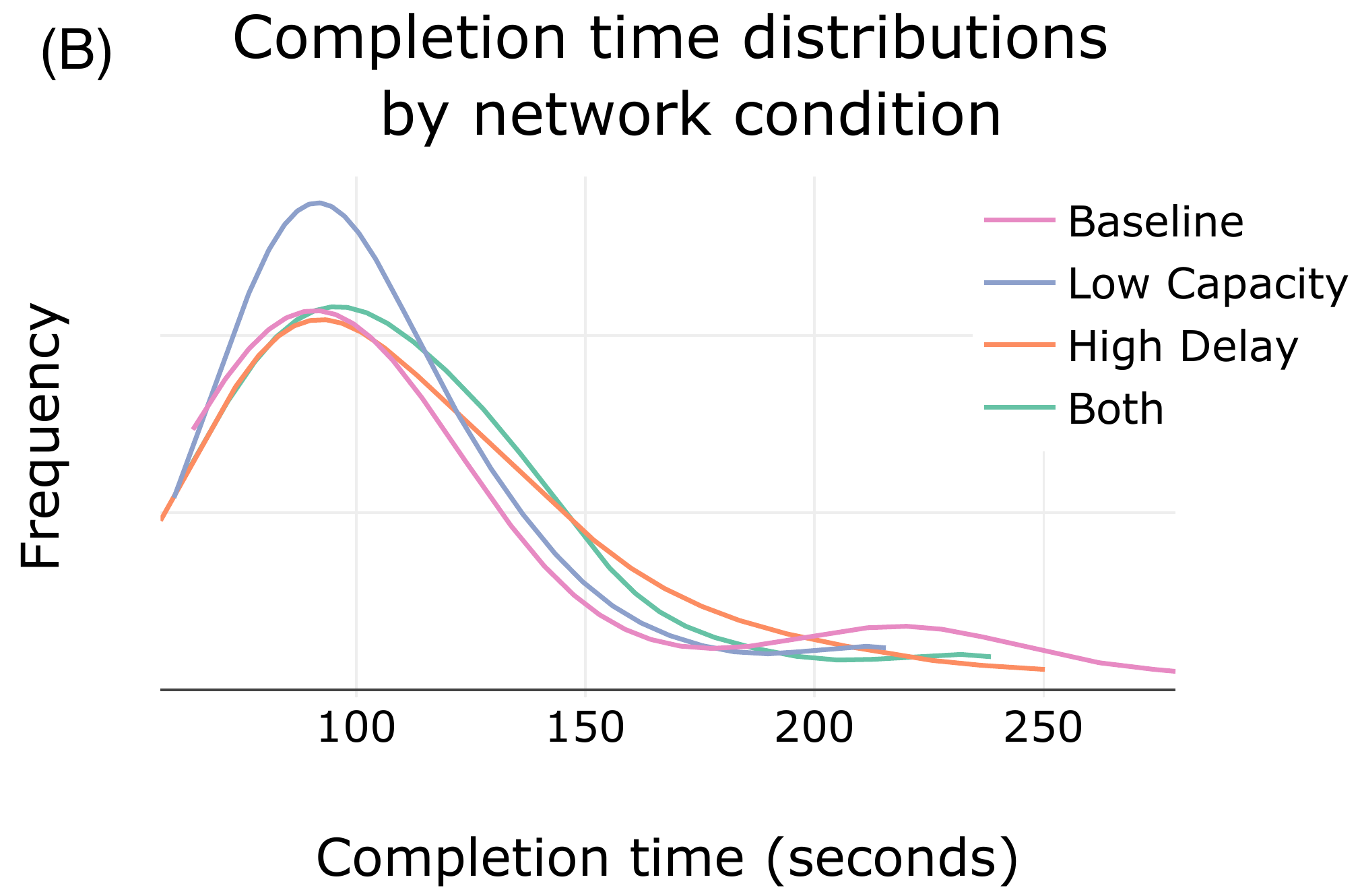}
\caption{\textbf{System Analysis}: \textbf{(A)} Comparison of completion time distributions of different user interfaces on the \textit{picking} task. \textbf{(B)} Comparison of completion time distributions of different network conditions using the phone interface for the \textit{picking} task.}
\label{fig:ui-hists}
\end{figure}



We compare all four user interfaces on both the \textit{lifting} task and the \textit{picking} task. Results on both tasks show that the keyboard interface was significantly slower than the other three.  Fig.~\ref{fig:ui-hists}A shows that the 3D mouse requires a mean completion time of 112.57 seconds on the \textit{picking} task, being slower than both the phone and VR controller. Please refer to Appendices~\ref{sec:uiLifting} and \ref{sec:uiDetails} for detailed data on completion times.  Fig.~\ref{fig:ui_diags}b summarizes the results of Kolmogorov-Smirnov tests over distribution of completion times for each interface on the \textit{picking} task. We observe that the keyboard (mean 151.45 seconds) and 3D mouse (mean 112.57 seconds) interfaces are statistically slower than the Phone~(mean 89.97 seconds) and VR~(mean 79.36 seconds) at a 5\% significance level. However, we were unable to conclude that the distribution of completion times on the Phone interface was different than that of the VR controller.
These results support our hypothesis that the \sysAbbr phone interface is similar in performance to the VR interface, but widely available owing to the ubiquity of the iPhone.


\subsection{Robustness to Network Capacity and Delays}\label{sec:network}
We evaluate the performance of \sysAbbr under various network conditions. We used the Cellsim network simulation tool~\cite{winstein13stochastic} to regulate the network capacity and delay between the platform and the users. Cellsim can faithfully replay network conditions recorded from real networks or operate according to synthetically generated network traces. Using Cellsim, we characterized the platform under four different network conditions. The first test was a baseline where the network had a constant 2.4Mbps of up/downlink bandwidth and 20ms of one-way delay. The second test emulated a low-capacity network link (constant 500Kbps; 20ms delay), the third emulated a high-delay link (constant 2.4Mbps; 120ms delay), and the last emulated a low-capacity and high-delay link (constant 500Kbps; 120ms delay). We used 500Kbps as a conservative estimate of a slow internet connection (such as a 3G network) and 2.4Mbps as an estimate of a slow home consumer connection. We used 120ms of one-way delay to approximate a Trans-Pacific connection (e.g. from the US to China) which was confirmed as a reasonable estimate through real world testing (see Sec. \ref{sec:pacific}). Please refer to Table~\ref{tab:ks-network-table} in the Appendix for further details on Kolmogorov-Smirnov tests over these distributions.

Evaluations performed on the three troublesome network traces produced roughly the same distributions of completions times as the baseline (see Fig.~\ref{fig:ui-hists}). This robustness evaluation of \sysAbbr{} under a wide range of network conditions is a first step to affirm its capacity to serve clients globally regardless of their locations. 



\begin{figure}[t]
\centering
\includegraphics[width=0.49\linewidth]{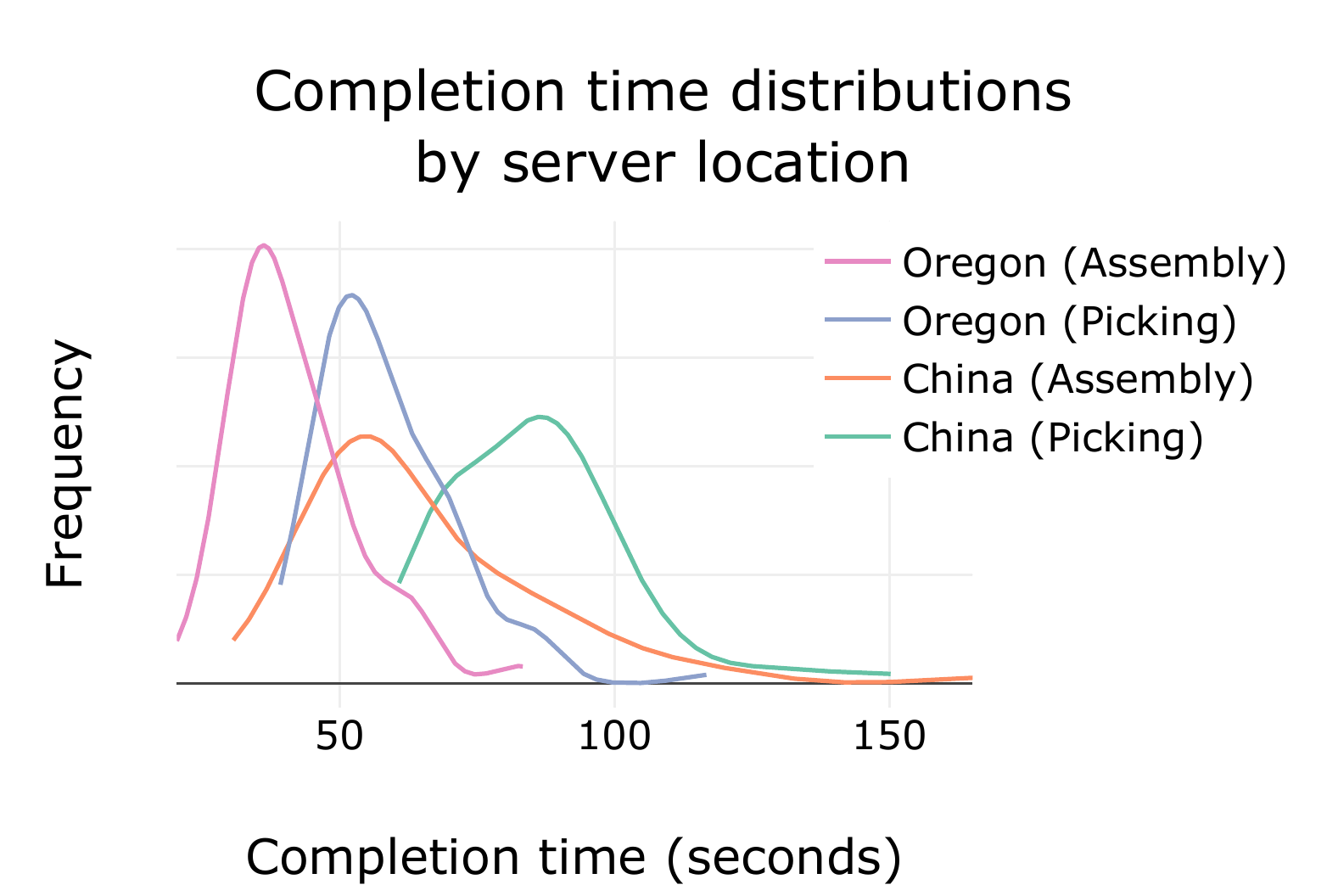}
\,
\includegraphics[width=0.48\linewidth]{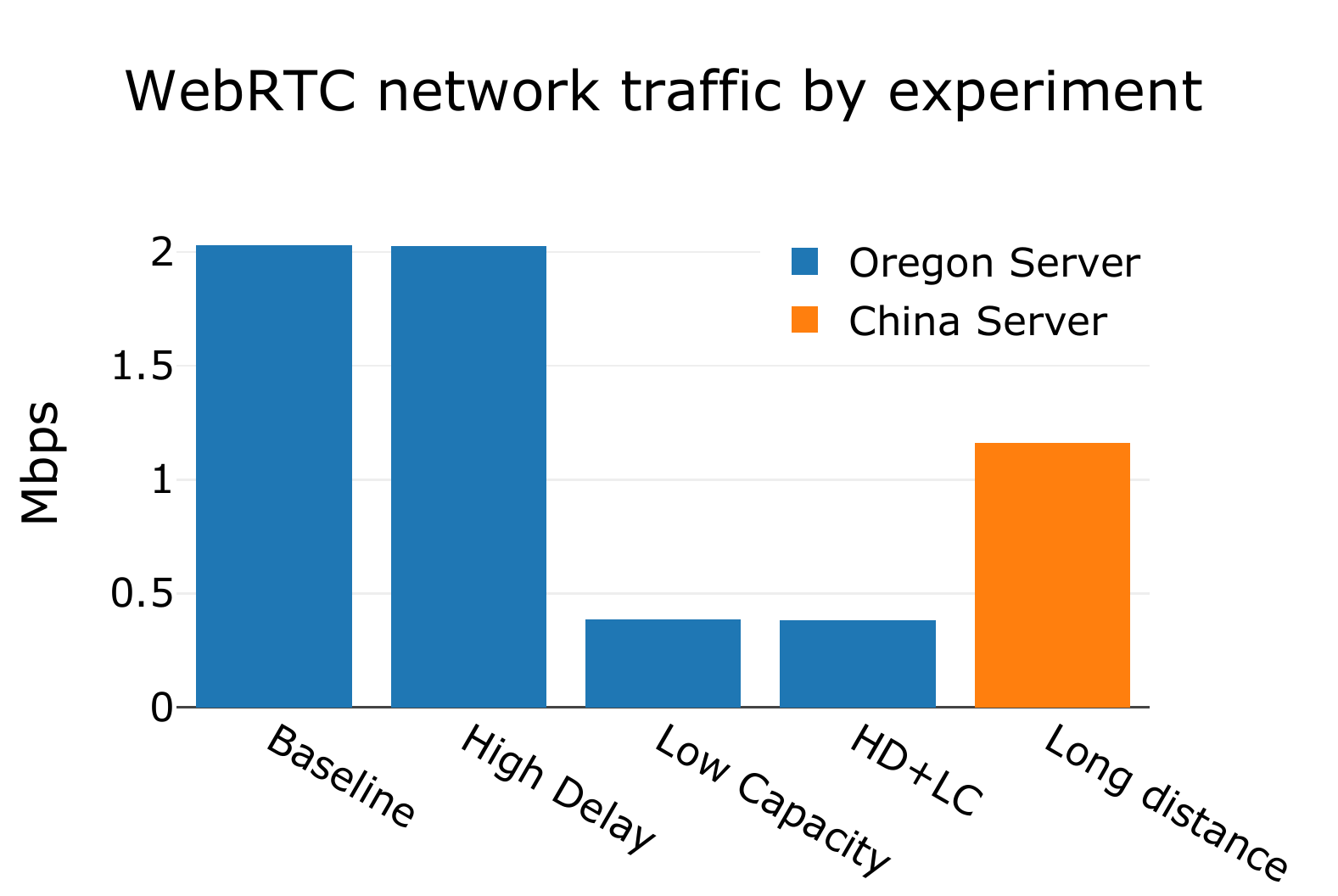}
\caption{\textbf{System Stress Test}: Left: Comparison of completion time distributions between servers located in Oregon vs. in China, when tested from California. Right: Comparison of average throughput of video stream by network experiment, as well as for cross-pacific experiments.}
\label{fig:pacific}
\end{figure}

\subsection{System Stress Test: Remote Teleoperation from California to China}\label{sec:pacific}

We also evaluate \sysAbbr for demonstration collection at large distances. We compare the platform's performance over real-world networks across different geographic locations. In the first experiment, users in California performed tasks using a deployment of \sysAbbr{} running in a data center in China (approximately 6500 miles away); in the second experiment, the same users performed the same tasks but with \sysAbbr{} running in a data center in Oregon (approximately 500 miles away). Qualitatively, the users found that using the data center in China introduced a nearly constant delay compared to the Oregon servers. Users were able to complete the tasks despite the high delays in network communication to and from China, but completion times were slower than when using the Oregon data center (Fig. \ref{fig:pacific}). The mean completion times different by 24 and 28 seconds for the \textit{assembly} and \textit{picking} tasks respectively. Overall, this stress test demonstrates that \sysAbbr is robust to poor real-world network conditions, allowing for tasks to be successfully completed (albeit at a slower pace) and has potential to connect servers to remote workers worldwide.









\section{Policy Learning from Teleoperated Demonstrations}

In this section, we evaluate the utility of the demonstration data collected by \Sys{}. To this end, we collected a pilot dataset on two challenging manipulation tasks, and we answer the following questions: (1) Can the collected demonstrations help enable, accelerate, and improve policy learning? (2) Is a larger number of task demonstrations valuable to learning? The second question is  vital, as the core goal of \Sys is to scale demonstration collection.


\subsection{Data Collection and Task Setup}
As an initial pilot for our platform, we collected a dataset of over 1000 demonstrations on both the \textit{picking} and \textit{assembly} tasks with 20 hours of total platform usage by using contractors that were located remotely. Indeed, this validates the notion that \Sys{} is a dynamic data collection platform that can be used to quickly crowdsource data on-demand. The Appendix contains some additional information and statistics on the dataset. 

We evaluate the utility of the collected demonstrations for learning policies with reinforcement learning from sparse rewards. We consider the \textit{Bin Picking (Can)} and \textit{Nut-and-peg Assembly (Round)} tasks from the Surreal Robotics Suite~\cite{fan2018surreal}, abbreviated as \textit{can picking} and \textit{round assembly} respectively. These are simplified versions of the full \textit{picking} and \textit{assembly} tasks, where the goal is to place a single can into its corresponding bin and to fit a single round nut onto its corresponding peg respectively. We do not use reward shaping; a non-zero reward is provided to the agent only if the agent successfully completes the task. This makes exploration challenging -- the agent has little chance of finding a goal state via random exploration due to the compositional structure of these tasks. 


\subsection{Demonstration-Guided Reinforcement Learning}

To leverage demonstrations during reinforcement learning, we initialize training episodes from states sampled randomly from the demonstration trajectories and do policy learning using a distributed implementation of proximal policy optimization (PPO)~\cite{schulman2017proximal, heess2017emergence}. This simple method for doing reinforcement learning with demonstrations has been motivated theoretically by \cite{kakade2002approximately} and used in practice in prior and concurrent work \cite{nair2017overcoming, resnick2018backplay, fan2018surreal}. 
We encourage using states visited in demos, as opposed to new experience by setting a myopic horizon of 100 timesteps for training episodes.


To investigate the benefits of leveraging more demonstrations to guide reinforcement learning, we compared utilizing None (pure RL), 1, 10, 100, and 1000 demonstrations on the two tasks. We trained policies from 10 different random seeds for each task and demonstration count and report the mean and standard deviation of the final policy return after 24 hours on the \textit{can picking} task and 48 hours on the \textit{round assembly} task. The results are presented in \tabref{demoresults} and additional details are described in Appendix~\ref{sec:learningDetails}.


\begin{table}[!t]
\centering
\begin{tabular}{lccccc}
& \multicolumn{5}{c}{\textbf{Number of Demonstrations}}   \\
\multicolumn{1}{l}{\textbf{Task}}                   & None & 1  & 10 & 100 & 1000 \\\hline
\multicolumn{1}{l}{\textbf{Bin Picking (Can)}}     & $0 \pm 0$ & $278 \pm 351$ & $273 \pm 417$ & $385 \pm 466$  & $\mathbf{641 \pm 421}$   \\
\multicolumn{1}{l}{\textbf{Nut-and-peg Assembly (Round)}} & $0 \pm 0$ & $381 \pm 467$ & $663 \pm 435$ & $575 \pm 470$  & $\mathbf{775 \pm 388}$   \\ \hline
\end{tabular}
\vspace{2mm}
\caption{\textbf{Quantitative results of learning with demonstrations.} This table shows how the average performance of trained policies varies with the number of demonstrations utilized during policy learning. The maximum possible return for each task is 1000 --- this would correspond to a policy that instantaneously solves the task. We observe that utilizing more demonstrations generally leads to better average task performance, and our best experimental results were obtained by using 1000 demonstrations (nearly the entire \Sys{} dataset) on each task.}
\label{tab:demoresults}
\end{table}

\subsection{Results}

Table~\ref{tab:demoresults} demonstrates the value of leveraging large numbers of demonstrations to guide reinforcement learning. Using 1000 demonstrations (nearly the entire \Sys{} pilot dataset) resulted in the best mean task performance on both tasks. This implies that leveraging larger numbers of demonstrations can help policies learn to solve the task more consistently. The results reported in Table~\ref{tab:demoresults} exhibit high variance since every setup had some fraction of the 10 different policies that were unable to solve the task successfully in the allotted time, and consequently had a return of 0. There are also policies that could not solve the task consistently from every start state, or policies that took longer to solve the task. Both behaviors result in lower average returns for the policy. Thus, the high average performance for the 1000 demonstration experiments suggest that the increased number of demonstrations enable consistently training near-optimal policies that solve the task. 

We observed a much higher gap in performance between the 1000 demonstration results and the other results for the \textit{can picking} task than for the \textit{round assembly} task. We suspect that this is because policies on the \textit{round assembly} task were given more time to train. We allowed a period of 48 hours for the \textit{round assembly} task because we found that 24 hours was an insufficient amount of time for policies to solve the task. The larger quantity of agent experience collected during the 48 hours could explain how some agents trained with only 10 demonstrations were able to achieve comparable performance to agents trained with 100 demonstrations. 
Over longer time periods, the benefits of experience diversity present in the 1000 demonstration dataset could plausibly be mirrored by using 10 demonstrations and giving our distributed agents enough time to explore from demonstration states. 
However, we emphasize that tasks that exhibit greater diversity between episodes or are compositionally complex (such as the full version of our simulated tasks) will extend this gap, needing more supervisory data. Similarly, policy learning methods that leverage the demonstrations in more sophisticated ways might also overcome these limitations. Nevertheless, our findings indicate that a large and diverse dataset of demonstrations can encourage agents to explore and learn more effectively.




\section{Conclusion}
\label{sec:conclusion}

We presented \Sys{}, a platform that supports large-scale on-demand data collection for robot learning tasks. We conducted a user study to evaluate the efficacy of our iPhone user interface and WebRTC-based communication backend, and also demonstrated that our platform is able to support real-time trans-Pacific communication and control without a significant difference in performance and task completion times. These results suggest that \Sys{} can be used by a significant portion of the worldwide population, regardless of geographic location.



As an initial pilot for \sysAbbr, we collected a large pilot dataset consisting of over 2200 total successful demonstrations in only 22 hours of total platform usage on the \textit{Bin Picking} and \textit{Nut-and-peg Assembly} tasks. We used demonstrations collected from our platform to train policies on the \textit{Bin Picking (Can)} and \textit{Nut-and-peg Assembly (Round)} tasks, which are simplified variants of the original tasks, showed that the data allows the policies to solve the tasks with reinforcement learning from sparse rewards, and found that performance generally improves with the quantity of demonstrations used. Our final model is a very simple way to utilize the demonstrations by controlling the start state distribution that the agent sees during training --- we envision that more sophisticated algorithms can better leverage the demonstrations we collected. Future work will involve utilizing the platform to collect more data on a larger collection of diverse tasks, extending the platform to support remote teleoperation of real robot arms, and developing more sophisticated algorithms that can leverage large quantities of data collected for policy learning. For additional results, videos, and to download our pilot dataset, visit \href{https://roboturk.stanford.edu}{\texttt{roboturk.stanford.edu}}.

\clearpage
\acknowledgments{
\footnotesize
We thank DataTang for supporting our data collection efforts and making it possible for us to collect our pilot dataset. We acknowledge the support of ONR - MURI (1175361-6-TDFEK), Toyota Research Institute (S-2015-09-Garg), Weichai America Corp.(1208338-1-GWMZY), CloudMinds Technology Inc. (1203462-1-GWMXN), Nvidia (1200225-1-GWMVU), Toyota (1186781-31-UBLFA), Panasonic, and Google Cloud. We would also like to thank members of the Stanford People, AI \& Robots (PAIR) group (\href{http://pair.stanford.edu}{\texttt{pair.stanford.edu}}) and the anonymous reviewers for their constructive feedback. We thank Pengda Liu for help with infrastructure.}

\renewcommand{\baselinestretch}{0.88} 
\renewcommand*{\bibfont}{\small}
\bibliography{ccr}

\renewcommand{\baselinestretch}{1} 
\appendix
\normalsize

\section{Appendix}

\subsection{Simulation Environment Details}
\seclabel{taskDetails}
\begin{figure}[ht]
    \includegraphics[width=\linewidth]{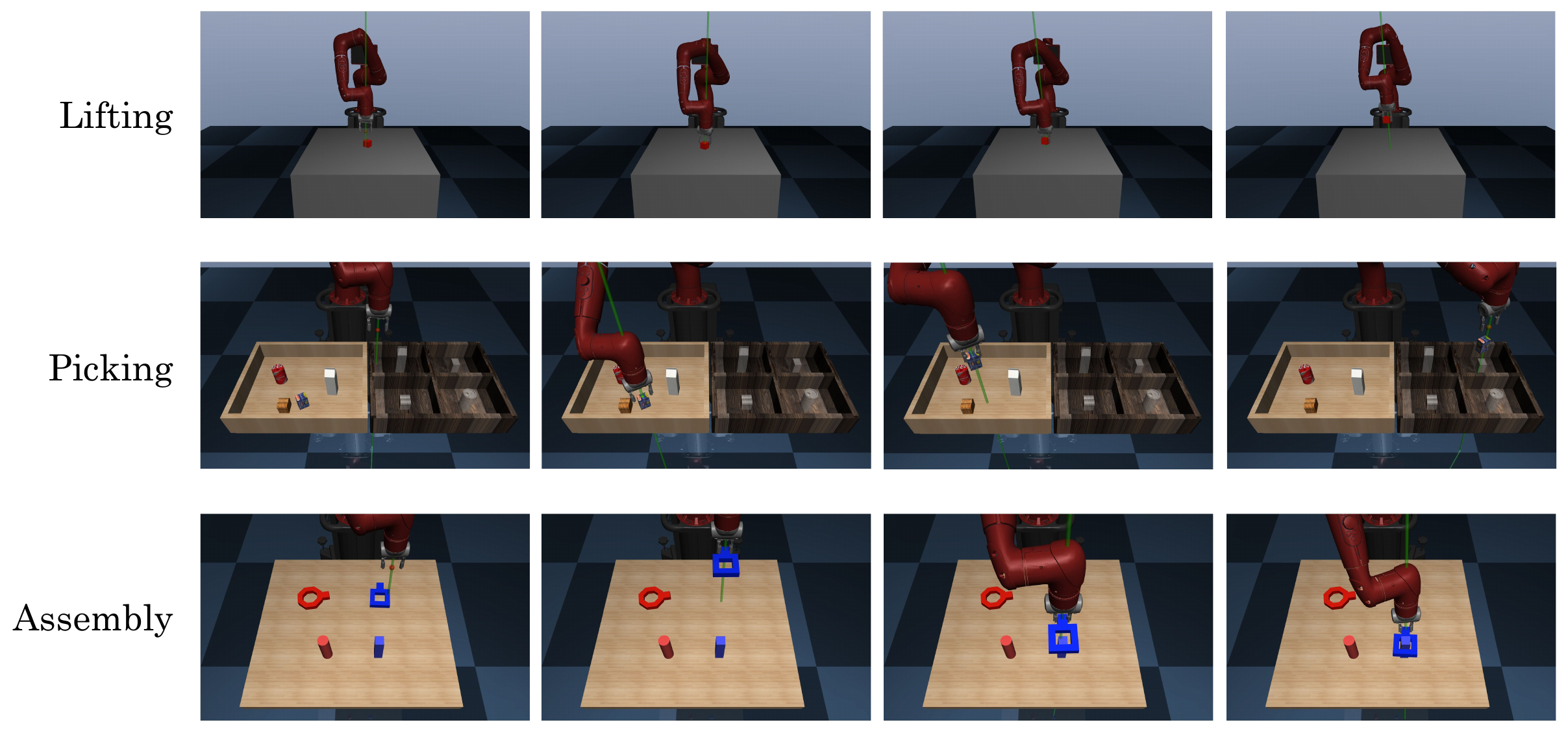}
    \caption{\textbf{Task Details}: A detailed depiction of the tasks used to evaluate our system. Each simulated manipulation environment contains a 7-DoF Sawyer robot arm and various objects. In the \textit{lifting} task (top row) the objective is to control the robot arm to grab the cube and lift it. In the \textit{picking} task (middle row) the objective is to place each object into its corresponding bin. In the \textit{assembly} task (bottom row), the objective is to place a round nut and square nut onto their corresponding pegs. These screenshots were taken along successful demonstrations of each task.}
    \label{fig:manip_tasks}
\end{figure}

\noindent \textbf{Environment design.}
In this work, we aim at demonstrating the effectiveness of our system in challenging multi-staged tasks which require both low-level dexterity and high-level reasoning. We evaluate our system in three simulated manipulation tasks with the Sawyer robot. The manipulation tasks were developed in the MuJoCo physics engine \cite{todorov2012mujoco}, which provides a high-speed simulation of contact dynamics with rigid 3D objects. The robot interacts with a set of objects in the workspace. Each operator observes the robot workspace from a camera facing the robot.

We describe the three tasks in Fig.~\ref{fig:manip_tasks}. First, we use a \textit{lifting} task as a diagnostic example to evaluate our choice of control interface and our system's characteristics. We also evaluate our system with two complex tasks: the \textit{picking} task and \textit{assembly} task. We perform large-scale data collection and model training on the \textit{picking} task and the \textit{assembly} task. These tasks involve a variety of dexterous manipulation skills, such as grasping, pick-and-place, nut assembly, and high-level perception tasks of understanding object categories and geometries. 

The above tasks resemble subtasks necessary for robot automation of part assembly. These tasks were inspired by the assembly challenge in the industrial robotics category of the World Robot Summit \cite{WorldRobotSummit}. The \textit{picking} task is similar to the kitting task, where parts need to be picked and laid out in preparation for assembly, and the \textit{assembly} task is similar to the task board task, where nuts must be placed onto designated pegs on a board. Fig.~\ref{fig:manip_tasks} shows screenshots taken along a successful demonstration of each of the three tasks. 

\subsection{UI Experiments: Lifting Task}
\seclabel{uiLifting}
\label{sec:analysis-lift}
\begin{figure}[th]
\centering
\includegraphics[width=.49\linewidth]{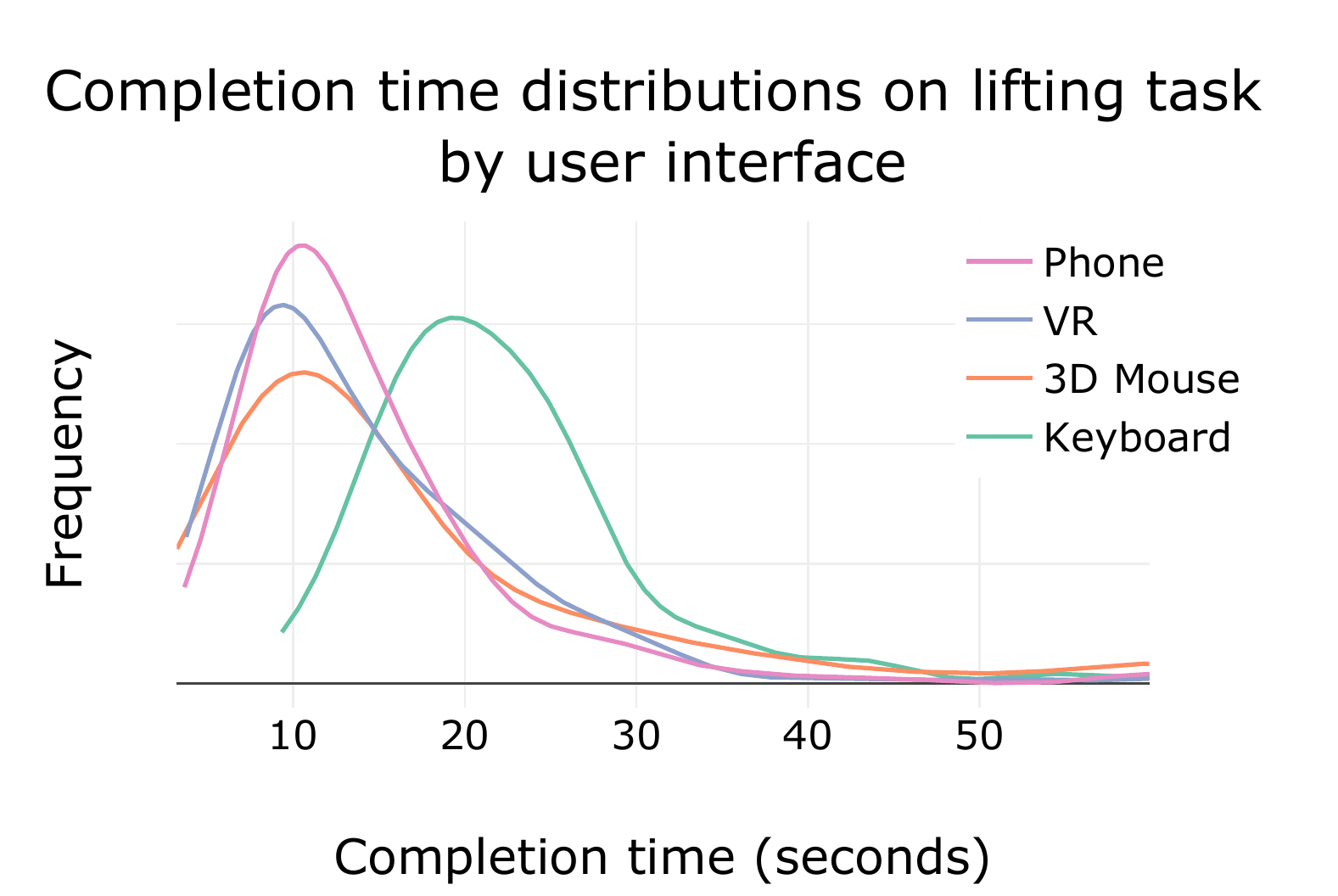}
\includegraphics[width=.49\linewidth]{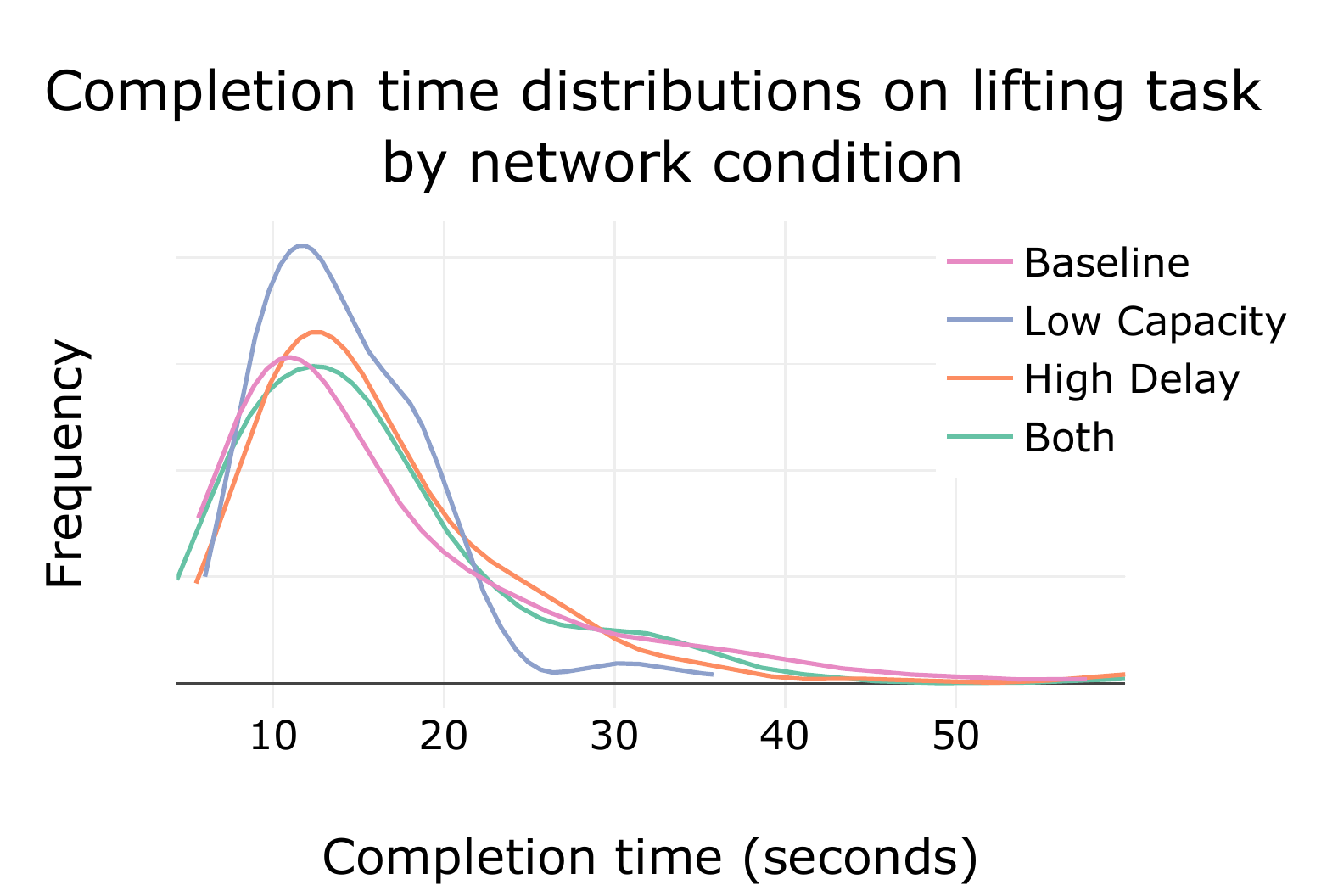}
\caption{Comparison of completion time distributions of different user interfaces on the \textit{lifting} task.}
\label{fig:ui_network_hists}
\end{figure}

In Sec.~\ref{sec:analysis} we described the UI and network experiments we performed. We ran experiments testing four different user interfaces on both the \textit{lifting} and \textit{picking} tasks. Here, we describe the results of these experiments on the \textit{lifting} task. As Fig.~\ref{fig:ui_network_hists} shows, users took significantly longer to complete the lifting task using the keyboard interface as compared to the other three interfaces, which all perform roughly equally. The lifting task is relatively simple and every episode of interaction is around 10 seconds, which was too short to reveal any significant differences between the phone, VR, and 3D mouse interfaces.

\subsection{UI Experiments: Details}
\seclabel{uiDetails}

In Sec.~\ref{sec:analysis} and Sec.~\ref{sec:analysis-lift} we described the general trends that we found during our UI experiments. In Table~\ref{tab:lift-stats} and Table~\ref{tab:pick-stats}, we detail the concrete results of those same experiments.

\begin{table}[p]
    \centering
    \caption{Statistics on completion times of the \textit{lifting} task across different user interfaces.}
    \begin{tabular}{l|c|c|c|}
        \cline{2-4}
        & Mean & Standard Deviation &  Num. Demonstrations  \\ \hline
        \multicolumn{1}{|l|}{Keyboard} & 22.34 & 7.92 & 187 \\ \hline
        \multicolumn{1}{|l|}{3D Mouse} & 16.58 & 12.27 & 193 \\ \hline
        \multicolumn{1}{|l|}{VR Controller} & 14.29 & 8.38 & 160 \\ \hline
        \multicolumn{1}{|l|}{Phone} & 14.26 & 8.27 & 167 \\ \hline
    \end{tabular}
    \label{tab:lift-stats}
\end{table}

\begin{table}[p]
    \centering
    \caption{Statistics on completion times of the \textit{picking} task across different user interfaces.}
    \begin{tabular}{l|c|c|c|}
        \cline{2-4}
        & Mean & Standard Deviation &  Num. Demonstrations  \\ \hline
        \multicolumn{1}{|l|}{Keyboard} & 151.45 & 23.69 & 40 \\ \hline
        \multicolumn{1}{|l|}{3D Mouse} & 112.57 & 43.65 & 40 \\ \hline
        \multicolumn{1}{|l|}{VR Controller} & 79.36 & 29.41 & 40 \\ \hline
        \multicolumn{1}{|l|}{Phone} & 89.97 & 34.94 & 40 \\ \hline
    \end{tabular}
    \label{tab:pick-stats}
\end{table}

\subsection{Network Experiments}

We ran tests on four different network conditions, with bandwidth at 2.4Mbps or 500kbps and delay at 20ms or 120ms. The \textit{baseline} condition had 2.4Mbps bandwidth and 20ms delay. The \textit{low capacity} condition had bandwidth reduced to 500kbps and the \textit{high delay} condition had delay increased to 120ms. The \textit{both} condition had the worst bandwidth and delay. Each user completed the \textit{lifting} and \textit{picking} task at varying network condition levels. 
Fig.~\ref{fig:ui_network_hists} shows how similar the completion time distributions were for the lifting task, and Table~\ref{tab:network-stats-lift} shows the same results with the mean and standard deviation for each distribution.

\begin{table}[p]
    \centering
    \caption{Statistics on completion times of the \textit{lifting} task across different network conditions.}
    \begin{tabular}{l|c|c|c|}
        \cline{2-4}
        & Mean & Standard Deviation &  Num. Demonstrations  \\ \hline
        \multicolumn{1}{|l|}{Baseline} & 15.66 & 8.22 & 160 \\ \hline
        \multicolumn{1}{|l|}{Low Capacity} & 16.13 & 9.41 & 168 \\ \hline
        \multicolumn{1}{|l|}{High Delay} & 14.19 & 5.07 & 164 \\ \hline
        \multicolumn{1}{|l|}{Both} & 16.31 & 8.29 & 160 \\ \hline
    \end{tabular}
    \label{tab:network-stats-lift}
\end{table}

Table~\ref{tab:ks-network-table} shows the Kolmogorov–Smirnov statistic between distributions of completion times for each of our network experiments. Based on a statistical significance level of 5\%, we found no statistically significant difference in performance between network conditions, despite worsened network conditions. This is largely due to WebRTC's adaptive video compression keeping quality high and delays at a minimum. Table~\ref{tab:network-stats} shows the mean and standard deviation for each distribution compared in Table~\ref{tab:ks-network-table}.

\begin{table}[p]
    \centering
    \caption{Statistics on completion times of the \textit{picking} task across different network conditions.}
    \begin{tabular}{l|c|c|c|}
        \cline{2-4}
        & Mean & Standard Deviation &  Num. Demonstrations  \\ \hline
        \multicolumn{1}{|l|}{Baseline} & 111.54 & 39.84 & 40 \\ \hline
        \multicolumn{1}{|l|}{Low Capacity} & 113.93 & 51.62 & 40 \\ \hline
        \multicolumn{1}{|l|}{High Delay} & 106.4236 & 34.72 & 40 \\ \hline
        \multicolumn{1}{|l|}{Both} & 113.85 & 41.81 & 41 \\ \hline
    \end{tabular}
    \label{tab:network-stats}
\end{table}

\begin{table}[p]
\centering
\caption{Kolmogorov-Smirnov statistic between distributions and p-value for completion times in the \textit{picking} task across the different network configurations.}
\label{tab:ks-network-table}
\begin{tabular}{l|c|c|c|c|}
    \cline{2-5}
        & Baseline & Low Capacity & High Delay & Both \\ \hline
    \multicolumn{1}{|l|}{\multirow{2}{*}{Baseline}} & \multirow{2}{*}{-} & 0.125 & 0.165 & 0.150 \\ 
    \multicolumn{1}{|l|}{} & & (0.893) & (0.598) & (0.724) \\ \hline
    \multicolumn{1}{|l|}{\multirow{2}{*}{Low Capacity}} & 0.125 & \multirow{2}{*}{-} &  0.165 & 0.150 \\
    \multicolumn{1}{|l|}{} & (0.893) & & (0.603)  & (0.724) \\ \hline
    \multicolumn{1}{|l|}{\multirow{2}{*}{High Delay}} & 0.165 & 0.165 & \multirow{2}{*}{-} & 0.077 \\ 
    \multicolumn{1}{|l|}{} & (0.598) & (0.603) & & (0.999) \\ \hline
    \multicolumn{1}{|l|}{\multirow{2}{*}{Both}} & 0.150 & 0.150 & 0.077 & \multirow{2}{*}{-} \\ 
    \multicolumn{1}{|l|}{} & (0.724) & (0.724) & (0.999) & \\ \hline
\end{tabular}

\end{table}

\subsection{Analysis of object difficulty on the Bin Picking task}

\begin{figure}[h]
\centering
  \includegraphics[width=\linewidth]{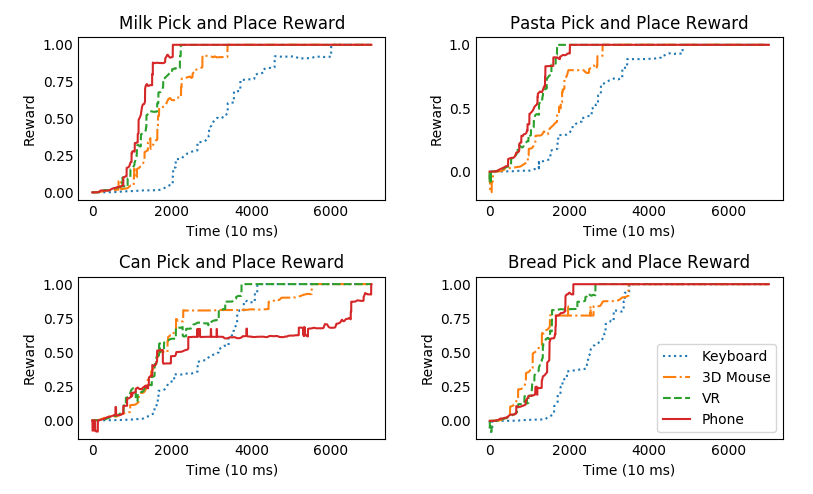}
  \caption{Comparison of environment rewards, which measure the completion of the pick and place operation.}
  \label{fig:ui_reward_comparison}
\end{figure}

\begin{figure}[h]
\centering
  \includegraphics[width=1.0\linewidth]{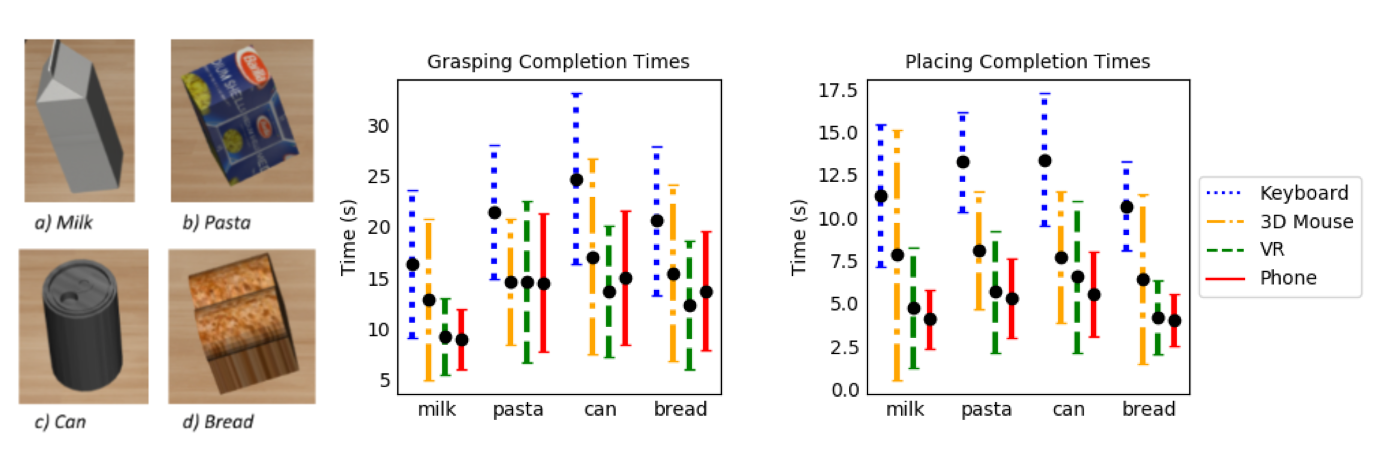}
  \caption{Time taken to grasp and place the four different types of objects in the pick and place task averaged across eight users for different user interfaces.}
  \label{fig:objects_UI}
\end{figure}

For the \textit{picking} task, we compared the time to grasp a target object and the time to place an object in a bin across the different user control interfaces.  Fig.~\ref{fig:ui_reward_comparison} shows a qualitative comparison of how users struggled with the various objects and interfaces. Fig.~\ref{fig:objects_UI} implies that milk tended to be the easiest object to grasp for all interfaces, while the rest of the objects were comparably difficult. Furthermore, we found that the phone experiments tended to have the most consistent results while the keyboard experiments showed a large increase in completion times.






\subsection{Teleoperation Controller Details}
\seclabel{sysDetails}


The teleoperation server maps received user phone poses and controls the simulated robot arm appropriately. Upon receiving a new phone pose with respect to the AR world coordinates of the phone, the controller matches this to a desired robot end effector pose. Next, an Inverse Kinematics procedure is invoked to compute a desired set point for the robot joint positions $\mathbf{q^*}$. These target joint positions are fed to a controller which computes joint velocities $\mathbf{\dot{q}} = -k_{v}(\mathbf{q} - \mathbf{q^*})$ and uses them to control the arm.

\subsection{Pilot Dataset Distribution}

Using \Sys{}, we were able to collect a total of $3224$ demonstrations over the course of $22$ hours of total system usage. A total of $2218$ demonstrations were successful and included in the dataest with $1171$ demonstrations on the \textit{picking} task and $1147$ on the \textit{assembly} task. These successful demonstrations correspond to $137.5$ hours of trajectories. 

\begin{figure}[t]
\centering
    \includegraphics[width=0.85\linewidth]{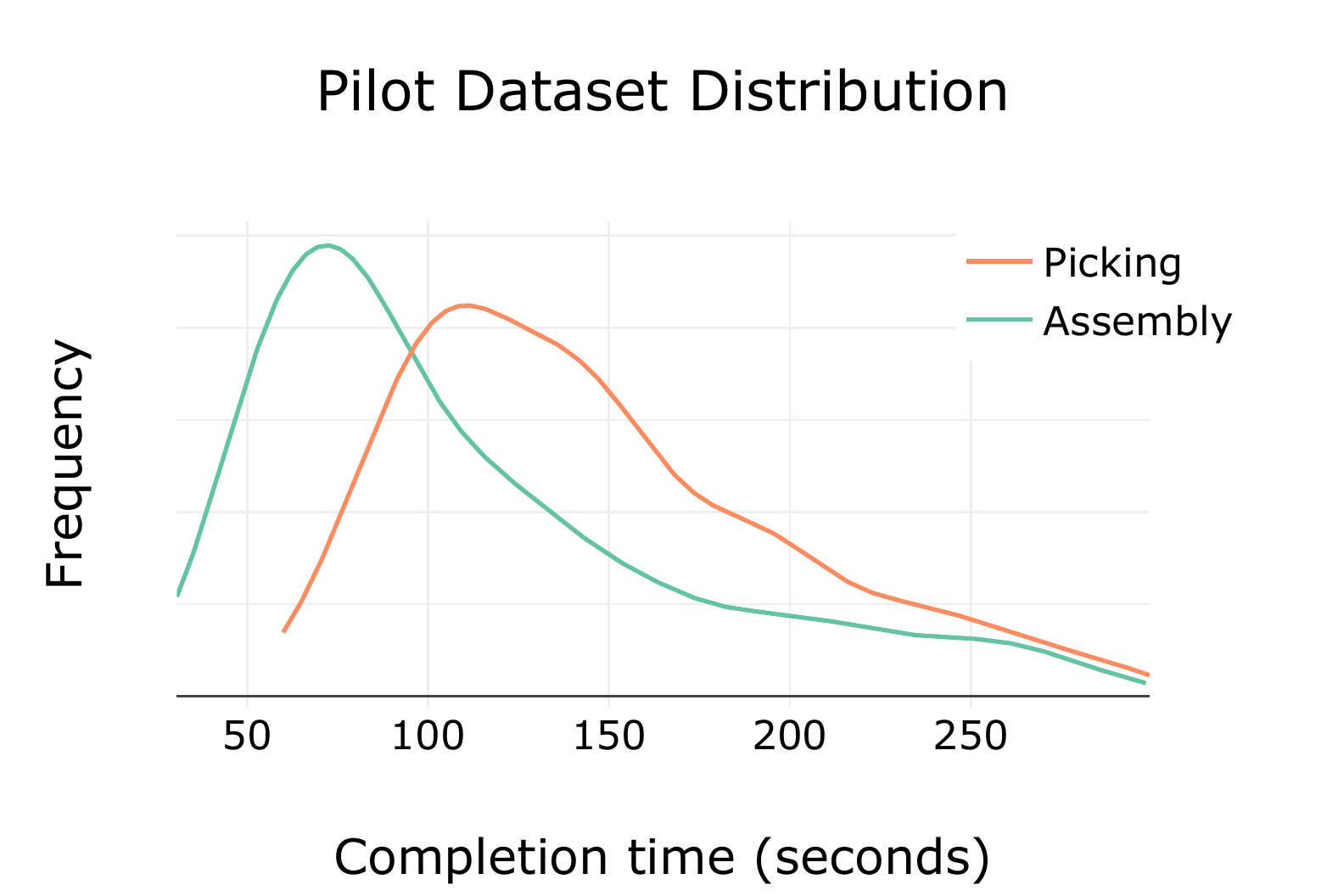}
    \caption{\textbf{\Sys{} Pilot Dataset Distribution}: Distribution of task completion times across all users in the large-scale dataset. 
    }
\label{fig:large_scale_completion}
\end{figure}

The mean completion time for the \textit{assembly} task is 112.12 seconds, with a standard deviation of 59.95. The mean completion time for the \textit{picking} task is 146.71 seconds, with a standard deviation of 52.67. The distribution of completion times in this dataset is shown in Fig.\ref{fig:large_scale_completion}.

\subsection{Policy Learning From Demonstrations}
\seclabel{learningDetails}

We ran our policy learning experiments on the \textit{can picking} and \textit{round assembly} tasks. For both tasks, the agent is given 100 time steps to solve the task, and is provided a sparse reward that is $1$ on task completion. 

We used the distributed PPO implementation from \cite{fan2018surreal} with 32 actors to train policies. We used neural networks with an LSTM layer of size 100 followed by two fully-connected layers of size 300 and 200 for both the actor network (mean) and the value function network. The policy is a Gaussian distribution parametrized by the mean network and a separate log standard deviation parameter. See \cite{fan2018surreal} for additional hyperparameter details. 

On every environment reset, with $90\%$ probability, a demonstration was sampled uniformly at random from the demonstration set, and the simulator was reset to a state sampled uniformly from the selected demonstration, and with $10\%$ probability, the environment was reset normally. 

Experiments on the \textit{can picking} task were each given 24 hours of training time, while those on the \textit{round assembly} task were given 48 hours of training time. These cutoff times were experimentally determined to be sufficient for training policies to complete the tasks.

\end{document}